\documentclass{article}
\usepackage{ijcai19}

\usepackage{times}
\usepackage{soul}
\usepackage{url}
\usepackage[utf8]{inputenc}
\usepackage[small]{caption}
\usepackage{graphicx}
\usepackage{amsmath}
\usepackage{booktabs}
\usepackage{algorithm}
\usepackage{algorithmic}

\usepackage{amssymb}
\usepackage{multicol}  
\usepackage{multirow}
\usepackage{float}

\title{Rethinking Loss Design for Large-scale 3D Shape Retrieval}

\author{
Zhaoqun Li$^1$\thanks{They contributed equally to this work}
\and
Cheng Xu$^1$\footnotemark[1] \and
Biao Leng$^{1,2,3}$\thanks{Corresponding author}
\affiliations
$^1$School of Computer Science and Engineering, Beihang University, Beijing, 100191\\
$^2$Beijing Advanced Innovation Center for Big Data and Brain Computing, Beihang  University, Beijing, 100191\\
$^3$State Key Laboratory of Virtual Reality Technology and Systems, Beihang  University, Beijing, 100191
\emails
\{lizhaoqun, cxu, lengbiao\}@buaa.edu.cn
}

\begin{document}

\maketitle

\begin{abstract}
Learning discriminative shape representations is a crucial issue for large-scale 3D shape retrieval.  
In this paper, we propose the Collaborative Inner Product Loss (CIP Loss) to obtain ideal shape embedding that discriminative among different categories and clustered within the same class. 
Utilizing simple inner product operation, CIP loss explicitly enforces the features of the same class to be clustered in a linear subspace, while inter-class subspaces are constrained to be at least orthogonal. 
Compared to previous metric loss functions, CIP loss could provide more clear geometric interpretation for the embedding than Euclidean margin, and is easy to implement without normalization operation referring to cosine margin. 
Moreover, our proposed loss term can combine with other commonly used loss functions and can be easily plugged into existing off-the-shelf architectures.
Extensive experiments conducted on the two public 3D object retrieval datasets, ModelNet and ShapeNetCore 55, 
demonstrate the effectiveness of our proposal, 
and our method has achieved state-of-the-art results on both datasets.
\end{abstract}


\begin{figure}[h]
\centering
\includegraphics[width=0.9\linewidth]{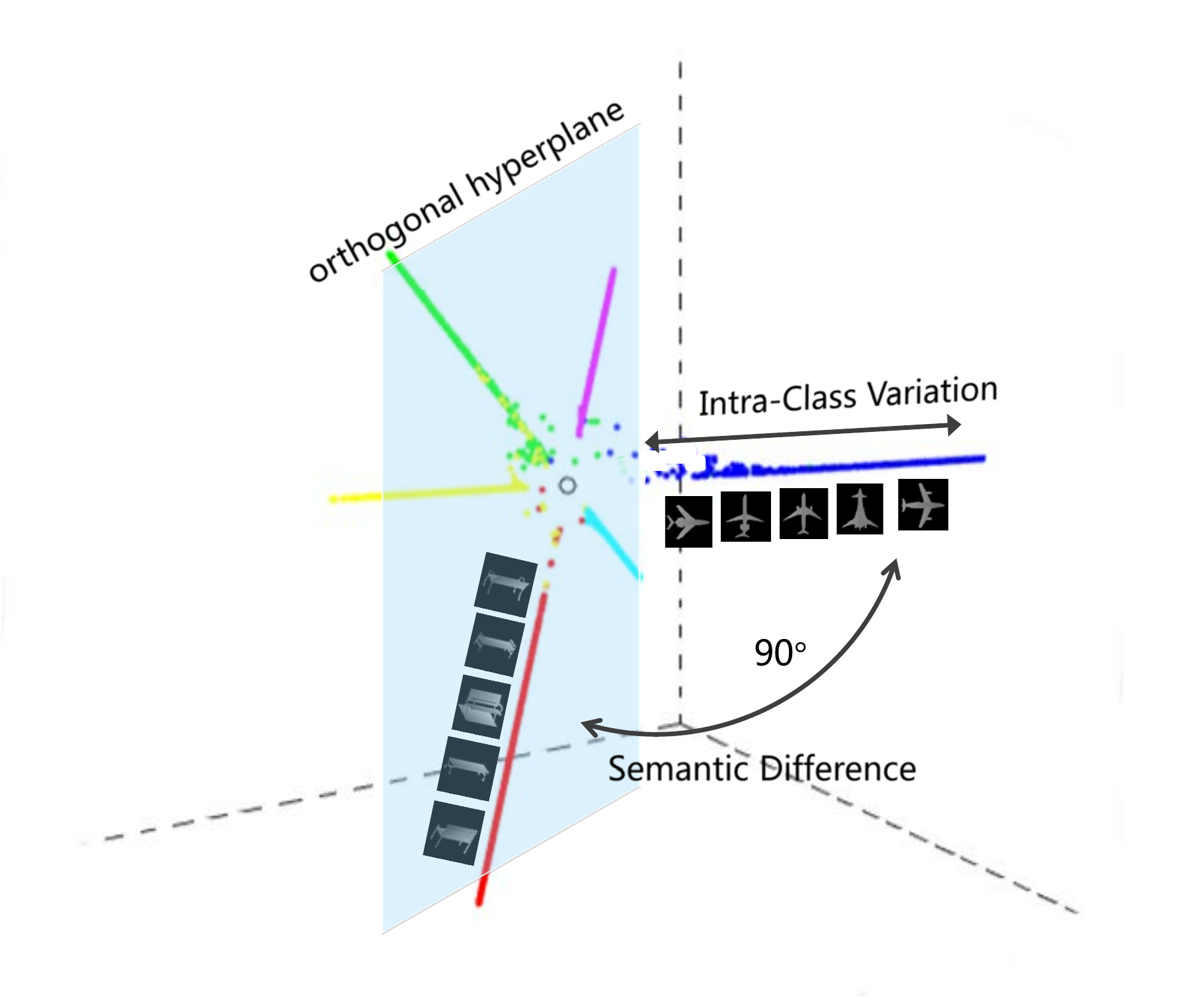}
\caption{Illustration of CIP loss effect in three-dimension. Different colors represent different classes which divide equally the whole space. 
In this distribution, the features of different classes are at least orthogonal and that of the same class are in a line. 
The norm of features corresponds to the intra-class variation and the angle corresponds to the semantic difference.}
\label{fig_ill}
\end{figure}

\section{Introduction}
3D shape retrieval is a fundamental problem in 3D shape analysis communities, spanning wide applications from medical imaging to robot navigation. 
With the development of Convolutional Neural Network (CNN) and the emergence of large-scale 3D repositories \cite{chang2015shapenet,wu20153d}, numerous approaches are proposed \cite{johns2016pairwise,su2015multi,xu2018emphasizing,li2019Angular} which significantly boost the performance in 3D shape retrieval task. Among these methods, view-based methods have achieved the best performance so far. 
In view-based 3D shape retrieval, images are first projected from different viewpoints of a 3D shape, and then they are passed into CNNs to obtain the discriminative and informative shape representation. The crucial issue of 3D shape retrieval is how to obtain ideal shape representations:
 \textbf{discriminative among different categories and clustered within the same class}.

The majority of view-based methods like MVCNN~\cite{su2015multi} train a CNN to extract shape representations with the standard softmax loss. 
While softmax loss does not explicitly enforce the distance between features in the embedding. 
To obtain ideal shape representations, ~\cite{he2018triplet} recently introduce deep metric loss functions to the 3D shape retrieval task. Such metric learning techniques, like center loss~\cite{10.1007/978-3-319-46478-7_31}, triplet loss~\cite{7298682} and triplet-center loss (TCL) \cite{he2018triplet}, are conducted in the Euclidean space and could remarkably boost the retrieval performance. 
However, these loss functions are designed by Euclidean margin, which is difficult to be determined because of the large span of Euclidean distance and has weak geometric constraint due to the variance of feature vector magnitude. 
With more clear geometric interpretation, metric loss functions based on cosine margin like coco loss~\cite{liu_2017_coco_v2} are particularly popular in the image retrieval and face verification communities. 
Considering the superior geometric constraint of cosine margin, we also introduce the coco loss to 3D shape retrieval in our experiment. 
The drawbacks with these methods lie in the complicated back-propagation calculation and the unstable parameter update, especially for large-scale 3D shape datasets. 
Therefore, we seek for a simple and stable metric loss with clear geometric interpretation to improve the discriminability of shape representations.

In this work, our intuition is that, for an ideal embedding space, the features of the same class should be in a line while at least orthogonal to other features.
Meanwhile, the simplicity and stability of the loss function must be taken into consideration in the loss design for large-scale dataset.
To this end, we propose a new Collaborative Inner Product Loss function (CIP Loss) to jointly optimize intra-class similarity and inter-class margin of the learned shape representations. 
In the loss design, we directly adopt the simple inner product which could provide distinct geometric constraints (see Fig.~\ref{fig_ill}) and enforce learned visual features $f$ to satisfy two conditions:
\begin{align}
&Dot(f_i,f_j) \rightarrow \infty \quad if \ y_i=y_j  \label{eq:rel1}\\
&Dot(f_i,f_j) \leq  0 \quad if \ y_i \neq y_j  \label{eq:rel2}
\end{align}
where $y$ is the label and $Dot(\cdot,\cdot)$ means the inner product between two vectors. 
On one hand, CIP loss encourages visual features of the same class to cluster in a linear subspace where the inner product between features tends to infinity, indicating the maximum extent of linear correlation among features. On the other hand, inter-class subspaces are requested to be at least orthogonal, meaning that for each category subspace, other category subspaces are squeezed to another half-space of its orthogonal hyperplane. 
In particular, if there are six categories in three-dimensional feature space, CIP loss would make all categories divide the whole embedding space equally as shown in Fig.~\ref{fig_ill}. Specially, in order to save computation, the inner product between visual features is replaced by the inner product between visual features and category centerlines in our method.

\begin{figure*}
\centering
\includegraphics[width=1\textwidth]{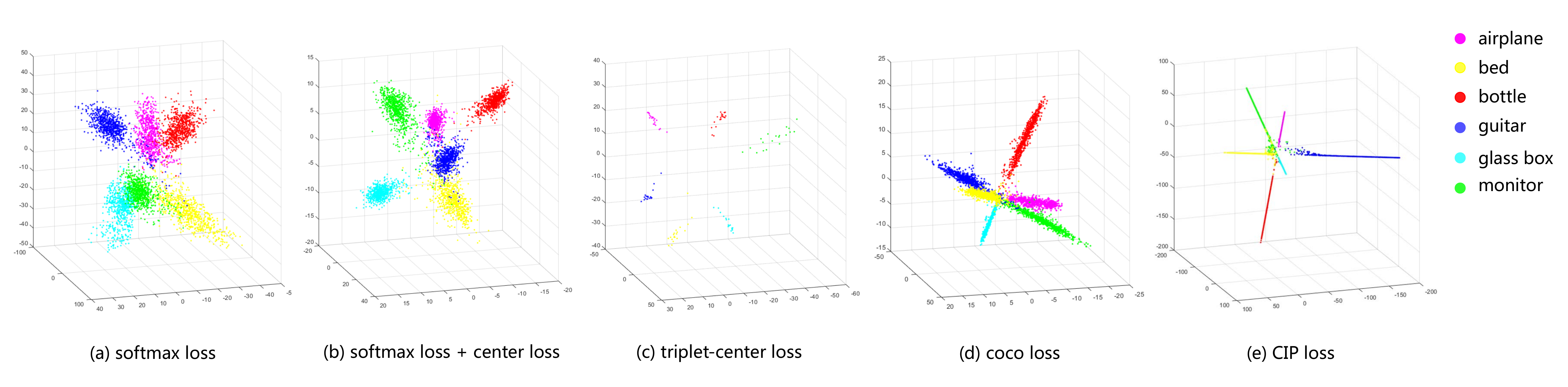}
\caption{Illustrative comparison between CIP loss and other loss functions on ModelNet test set. In three dimension, CIP loss can converge with at most 6 classes because Ortho loss constrains the features of different classes to be at least orthogonal.
So we select 6 classes from ModelNet and train CNNs on the corresponding train set.  
We show all the view features in (a)(b)(d)(e). For (c), only model features are available since triplet-center loss adopts MVCNN architecture.
Compared with other loss functions, CIP loss imposes the embedding space a more strict geometric constraint that makes the features orthogonal between different classes and clustered in a linear subspace with the same class.}
\label{fig_featuredistribution}
\end{figure*}

We show in Fig. \ref{fig_featuredistribution} the effect of different loss functions on the feature distribution.
Compared with previous metric learning techniques in 3D shape retrieval community, the proposed method has the following key advantages. 
1) CIP loss could provide more explicit geometric interpretation for the embedding than approaches based on Euclidean distance, see Fig. \ref{fig_featuredistribution}(b)(c)(e). 
2) Compared with popular cosine margin, CIP loss is simple and elegant to implement without margin design and normalization operation, which can stabilize the optimization process.
3) CIP loss is flexible to be plugged into existing off-the-shelf architectures, where it can work standalone or in combination with other loss functions.

In summary, our main contributions are as follows.
\begin{itemize}
\item We propose a novel metric loss function, namely collaborative inner product loss (CIP Loss), which adopts elegant inner product between features to perform more explicit geometric constraints on the shape embedding.
\item Two components of CIP loss are proposed, namely Cluster loss and Ortho loss. Cluster loss guarantees visual features of the same class to cluster in a linear subspace, while Ortho loss enforces inter-class subspaces are at least orthogonal, making shape representations more discriminative.
\item Our method achieves the state-of-the-art in the large-scale datasets, ModelNet and ShapeNetCore55, showing the effectiveness of CIP loss.
\end{itemize}

\section{Related work}
In this section, we briefly review some view-based approaches of 3D shape retrieval and introduce some typical deep metric learning methods in this domain.  

\subsection{Multi-view 3D shape retrieval}
3D shape retrieval methods could be roughly divided into two categories: model-based methods and view-based methods. Although the model-based methods~\cite{xie2017deepshape,qi2017pointnet++,shen2018mining} can capture the 3D structure of original shape by various data form, their performances are typically lower than view-based methods due to the complexity of computation and the noise in shape representation. 

The view-based 3D shape retrieval methods use rendered images to represent the original shape. MVCNN \cite{su2015multi} is a typical method adopting CNN to aggregate rendered images. In this method, the rendered images from different views are pooled to generate the shape feature. \cite{huang2018learning} adopts a local MVCNN shape descriptors, which use a local shape descriptor of each point on the shape to analyze the shape. Recently proposed GVCNN \cite{feng2018gvcnn} separates views in groups by intrinsic hierarchical correlation and discriminability among them, which largely improves the performance on the 3D shape retrieval. Another technique used in 3D shape retrieval is re-ranking. GIFT \cite{bai2016gift} adopts CNN with GPU acceleration to extract each 
single view feature and proposes the inverted file to reduce computation for fast retrieval. 

The aforementioned methods are trained under the supervision of softmax loss which is not always consistent with retrieval task. \cite{he2018triplet} recently puts forward TCL that combines the center loss and triplet loss, which achieve state-of-the-art results on various datasets. This shows that deep metric learning plays an important role in the 3D retrieval task.

\subsection{Deep metric learning}
As a key part of deep learning framework, loss design has been studied widely in retrieval and other domains.
Most commonly used loss functions in 3D shape retrieval are designed in Euclidean space, such as triplet loss, center loss and TCL.
Triplet loss \cite{7298682} forces inter-class distance exceed intra-class distance by a positive Euclidean margin which is widely applied in face recognition. 
However, the original triplet loss has the problem in computation cost that the number of triplets grows cubically with the dataset.
To tackle this problem, many improved version \cite{Song2016Deep,Song2016Deep} are proposed in various domains.
\cite{Song2016Deep} is an efficient approach that proposes an algorithm for taking full advantage of the training samples by using the matrix of pairwise distances and achieves high performance on image retrieval.
Another popular metric loss is center loss \cite{wen2016discriminative} that tries to gather the same class features in one point which is called center of the class.
Combining the advantages of triplet loss and center loss, TCL \cite{he2018triplet} is proposed for 3D shape retrieval which gains a better performance than the other loss functions.
However, they adopt the Euclidean margin which is difficult to design and has a weak geometric constraint.

On the other hand, cosine margin is recently proposed which are popular in face recognition. 
As a typical method, coco loss \cite{liu_2017_coco_v2} advocates that the weakness of softmax loss is in the final classification layer. 
By weight normalization and feature normalization, it introduces cosine margin to reduce intra-class variation.
Although the cosine margin has more clear geometric interpretation than Euclidean margin, it also brings instability in the optimization.
Inspired by these works, we design our loss function adopting inner product which is stable in the training process and efficient in computation.

\section{Methods}

Deep metric learning aims to train a neural network, denoted by $f_{\theta}(\cdot)$, 
which maps the original data onto a new embedding space $\mathbb{F} \in \mathbb{R}^n $. 
For writing convenience, we use $f_i \in \mathbb{R}^n$ to represent $f_{\theta}(x_i)$ which is the extracted feature from CNN. 
In this work, our goal is to make $\mathbb{F}$ ideal, which means that the intra-class angular distance is 0 while the inter-class angular distance is at least $\frac{\pi}{2}$.
To fulfill this goal, the loss design needs to consider two necessary tasks:
\emph{(1) Enlarging the inter-class distances between features} and \emph{(2) Reducing the intra-class distances between features}.
Then, the CIP loss could be formalized as the sum of pull term and push term corresponding to Cluster loss and Ortho loss respectively:
\begin{equation}
    L _{CIP}= L_{cluster} + \lambda \cdot L_{ortho} 
    \label{equation_loss}
\end{equation}
where $\lambda$ is a trade-off hyperparameter.

To achieve the aforementioned goal, a natural idea is adopting cosine distance which needs normalization of feature vectors, while this operation would bring complexity in backpropagation and instability in optimization that we will discuss in Sec. \ref{section_discussion}. 
In this article, we use the inner product of two vectors as the pairwise similarity for the sake of simplicity and stability:
\begin{equation}
  s(f_i, f_j) = f_i^\mathsf{T} \cdot f_j 
\end{equation}

Based on inner product similarity, we propose our Cluster loss and Ortho loss.

\subsection{Proposed loss functions}
Given a training dataset $ \mathcal{X} \in \mathbb{R}^D $, let $ \mathcal{Y}=\{ y_1,y_2,...,y_{| \mathcal{X} |} \} $ denote the corresponding label set,
$y_i \in \{1,2,...,K\} $ where $K$ is the number of classes. 

The Cluster loss constrains the same class features in a linear subspace, \emph{i.e.} a line in $\mathbb{F}$.
Inspired by center loss, we define this line, denoted by $c_k \in \mathbb{R}^n$, as the centerline of the $k^{th}$ class. 
Different from the interpretation of ``center'' in center loss,
the centerline here represents the direction of the corresponding class features.
Given a batch of training data with $M$ samples,
the pull term named Cluster loss is defined as:
\begin{equation}
    L_{cluster} =  \sum_{i=1}^{M} \frac{1}{f_i^\mathsf{T} \cdot c_{y_i} + d}  
    \label{equation_cluster}
\end{equation}
where $d=2$ is a constant value for numerical stability.
Cluster loss encourages $f_i$ to pull its corresponding centerline $c_{y_i}$.
And in turn, it forces the centerline to concentrate its corresponding features.

The design of Ortho loss considers that the different class features are at least orthogonal, according to condition~\ref{eq:rel2}.
As centerline represents the direction of the corresponding class, 
we penalize the features which are not orthogonal to other negative centerlines.
The push term named Ortho loss is formulated as:
\begin{equation}
    L_{ortho} = \sum_{i=1}^{M} \left[ \sum_{k=1, k \neq y_i}^{K} \max (f_i^\mathsf{T} \cdot c_k, 0) \right] 
    \label{equation_ortho}
\end{equation}
Ortho loss pushes $f_i$ to be at least orthogonal to all other centers so that the inter-class distance will increase.
The generated feature distribution of our method is shown in Fig. \ref{fig_featuredistribution}(e).

\subsection{Backpropagation}
\textbf{Gradient of feature vector.}
Owing to the simple formulation of loss functions and adoption of the inner product, 
the formulas of gradient for feature are also simple.
For writing convenience, we denote $(\cdot)_{+} = \max(\cdot,0)$. 
We use $\delta(cdt)=1$ to indicate $cdt$ is true and $\delta(cdt)=0$ otherwise.
The gradient of $f_i$ for $L_{ortho}$ is foumulated as follows:

\begin{equation}
\begin{aligned}
    \frac{\partial L_{ortho}}{\partial f_i} = \sum_{k=1,k \neq y_i}^{K} \delta(f_i^\mathsf{T} \cdot c_k > 0) \cdot c_k 
    \label{equation_bportho} 
\end{aligned}
\end{equation}

As for $L_{cluster}$, the existence of $d>0$ is to adjust the gradient of function $\frac{1}{x}$ which explodes on $x=0$.
We set $d=2$ in our implementation.
However the gradient function still has this problem when $f_i^\mathsf{T} \cdot c_{y_i} $ is near to $-d$.
In fact, the original formulation is:
\begin{equation}
\begin{aligned}
    \left( \frac{\partial L_{cluster}}{\partial f_i} \right)_{origin}
    = -\frac{1}{(f_i^\mathsf{T} \cdot c_{y_i} + d)^2} \cdot c_{y_i}
    \label{equation_bpcluster}
\end{aligned}
\end{equation}

Eq. \ref{equation_bpcluster} becomes $\infty$ when $f_i^\mathsf{T} \cdot c_{y_i}$ tends to be $-d$.
Thus we clip the term $f_i^\mathsf{T} \cdot c_{y_i}$ by 0 so that the optimization becomes stable.
Concretely, we use a surrogate gradient in backpropagation:
\begin{equation}
\begin{aligned}
    \left( \frac{\partial L_{cluster}}{\partial f_i} \right)_{srg}
    & = -\frac{c_{y_i}}{\left( (f_i^\mathsf{T} \cdot c_{y_i})_{+} + d \right) ^2}  \\
    \label{equation_bpclusters} .
\end{aligned}
\end{equation}

\noindent \textbf{Gradient of centerline.}
The gradient form is similar to that of feature vector since the inner product is a symmetric function.
In the same way, we adopt a surrogate gradient to evade gradient explosion for $L_{cluster}.$
\begin{equation}
\begin{aligned}
    \left( \frac{\partial L_{cluster}}{\partial c_i} \right)_{srg}
    = \sum_{j=1}^{M} -\frac{ \delta(y_j=i) \cdot f_{j}}{\left( (f_j^\mathsf{T} \cdot c_i)_{+} + d \right) ^2}
    \label{equation_bpcenters1} 
\end{aligned}
\end{equation}

For $L_{ortho}$, we replace the gradient of $L_{ortho}$ with respect to $c_i$ by a ``average'' version which can further stablize the update of center:

\begin{equation}
\begin{aligned}
   \left( \frac{\partial L_{ortho}}{\partial c_i} \right)_{srg} 
    = \frac{\sum_{j=1}^{M} \delta(y_j \neq i) \cdot \delta(f_j^\mathsf{T} \cdot c_i > 0) \cdot f_j}
    {1 + \sum_{j=1}^{M} \delta(y_j \neq i) \cdot \delta(f_j^\mathsf{T} \cdot c_i > 0)}
   \label{equation_bpcenters2} 
\end{aligned}
\end{equation}

\subsection{Discussion}
\label{section_discussion}

The loss function is essential for guiding the optimization that has a direct influence on the feature distribution. 
And in the training process, the convergence and the stability should be guaranteed, which poses challenges to the loss design. 
In this section, we will illustrate the motivation and the procedure of loss design in detail.

\noindent\textbf{Inner product without normalization}. 
In a typical CNN, the similarity between a deep feature vector $f_i$ and a layer weight $w_j$ is encoded by the inner product:
$s = f_i^\mathsf{T} \cdot w_j = ||f_i||_2  || w_j||_2 \cdot \cos(f_i,w_j) $.
In order to eliminate the effect of norms on similarity, some methods adopt normalization operations.
However, weight normalization leads to instability. We can infer from the gradient of weight after normalization:

\begin{equation}
    \frac{\partial}{\partial w} (\frac{w^\mathsf{T} \cdot f}{||w||_2})
    = \frac{f}{||w||_2} - \frac{(w^\mathsf{T} \cdot f) \cdot w}{||w||_2^3} 
    \label{equation_bpnormw}
\end{equation}

From Eq. \ref{equation_bpnormw} we can see that small value of $w$ will lead to gradient explosion.
In comparision, for the inner product, the gradient is $\frac{\partial (w^\mathsf{T} \cdot f)}{\partial w} = f$ which is more stable and computationally efficient.
So we choose the inner product for the sake of stabilization in the training process.
As for leaving out feature normalization, we take the same consideration.

In addition, this form doesn't need margin design or, from another perspective, we can regard the angular margin as 0 and $\frac{\pi}{2}$ in two terms corresponding that the inner product is $\infty$ and 0.

\noindent\textbf{Combination with other loss functions.}
Both of Cluster loss and Ortho loss can combine solely with softmax loss and bring improvement on performance.
It is worth pointing out that Ortho loss cannot maintain the centerlines solely since it focuses on enlarging inter-class distance.
When we combine $L_{ortho}$ with other loss functions, the centerlines will diverge.
In this case, we employ the batch version:
\begin{equation}
    L_{ortho,batch} = \sum_{i=1}^{M} \left[ \sum_{j=1, y_j \neq y_i}^{M} \max (f_i^\mathsf{T} \cdot f_j, 0) \right] 
    \label{equation_orthob}
\end{equation}

This batch version directly conducts the optimization on feature distribution without centerlines.
The backpropagation formula of batch version is similar to Eq. \ref{equation_bportho} that we only need to replace the role of $c_k$ by $f_j$.
We will discuss the performance difference of these two versions in the Sec. \ref{section_comploss}.

%

\section{Experiment}
\label{section_experiment}

\begin{figure*}
\centering
\includegraphics[width=\textwidth]{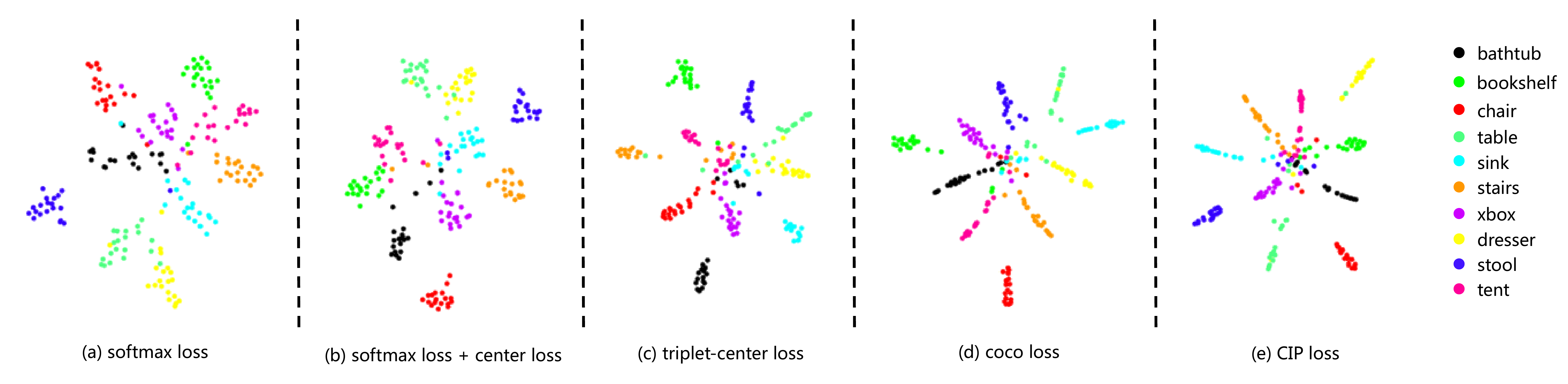}
\caption{Visualization of experiment results on ModelNet40. We randomly select 10 classes from the dataset and use t-SNE to reduce the dimension of shape features for visualization. }
\label{fig_sne}
\end{figure*}

In this section, we evaluate the performance of our method on two representative large-scale 3D shape datasets.
We first compare the results with the state-of-art methods. 
We also provide massive experiments to discuss the comparison and the combination with other metric loss functions. 
In the last part, we investigate the influence of the hyper-parameters and visualize the experiment results by dimension reduction.

\subsection{Retrieval on large-scale 3D datasets}

\begin{table}
\begin{center}
\scalebox{1}[1]{
\begin{tabular}{lccc}
\toprule[2pt]
Methods									& & AUC 			& MAP\\
\toprule[1.5pt]
ShapeNets 									& & 49.94\% 		& 49.23\% 		\\	
DeepPano 									& & 77.63\% 		& 76.81\% 		\\
MVCNN 									& & - 				& 80.20\% 		\\
GIFT 										& & 83.10\% 		& 81.94\% 		\\
Siamese CNN-BiLSTM					& & -				& 83.30\%			\\
PANORAMA-NN							& & 87.39\%		& 83.45\%			\\
GVCNN 									& & - 				& 85.70\% 		\\
RED										& & 87.03\% 		& 86.30\% 		\\
TCL										& & 87.60\% 		& 86.70\% 		\\
\hline
$L_{CIP}$								& & 87.56\% 			& 86.49\% \\
$L_{CIP}$ + softmax loss				& & 88.07\% 			& 87.08\%\\
$L_{CIP}$ + center loss					& & \textbf{88.21\%} 	& \textbf{87.22\%} \\
\bottomrule[2pt]
\end{tabular}
}
\end{center}
\caption{The performance comparison with state-of-the-arts on ModelNet40.}
\label{tab_sr}
\end{table}

\noindent\textbf{Implementation detail.} 
The experiments are conducted on Nvidia GTX1080Ti GPU and our methods are implemented by Caffe.
For the structure of the CNN, we use VGG-M \cite{chatfield2014return} pre-trained on ImageNet \cite{deng2009imagenet} as the base network in all our experiments. 
We use the stochastic gradient descent (SGD) algorithm with momentum 2e-4 to optimize the loss and the batch size is 100.
The initial learning rate is 0.01 and is divided by 5 at the 20th epoch. 
The total training epoch is 30.
The centerlines are initialized by a Gaussian distribution of mean value 0 and standard deviation 0.01.

Firstly, we render 36 views of each model by Phong reflection to generate depth images which compose our image dataset. 
The size of each image is 224x224 pixels in our experiment.
Then, we randomly select images from the image dataset to train our CNN.
In the test phase, the features of the penultimate layer, \emph{i.e. fc7}, are extracted for the evaluation.
We obtain the shape feature vector by averaging all view features of this shape. 
The cosine distance is adopted as the evaluation metric.

\noindent\textbf{Dataset.} 
We select two representative datasets, ModelNet and ShapeNetCore55, to conduct the evaluation of our methods. 
\emph{1)} ModelNet Dataset: this dataset is a large-scale 3D CAD model dataset composed of 127,915 3D CAD models from 662 categories. 
ModelNet40 and ModelNet10 are two subsets which contain 40 categories and 10 categories respectively.
In our experiment, we follow the training and testing split as mentioned in ~\cite{wu20153d}. 
\emph{2)} ShapeNetCore55: this dataset from SHape REtrieval Contest 2016~\cite{savva2016shrec} is composed of 51,190 3D shapes from 55 common categories divided into 204 sub-categories. 
We follow the official training and testing split to conduct our experiment.

The evaluation metrics used in this paper include mean average precision (MAP), area under curve (AUC), F-measure (F1) and normalized discounted cumulative gain (NDCG). 
Refer to~\cite{wu20153d,savva2016shrec} for their detailed definitions.

\begin{table}
\centering
\scalebox{0.88}[1]{
\begin{tabular}{lcccccc}
\toprule[2pt]
\multirow{2}{*}{Methods} & \multicolumn{3}{c}{Micro} & \multicolumn{3}{c}{Macro}\\
\cmidrule(lr){2-4}  \cmidrule(lr){5-7}  & F1 & MAP & NDCG & F1 & MAP & NDCG \\
\toprule[1.5pt]
Wang 			& 24.6 	& 60.0  	& 77.6 	& 16.3 	& 47.8 	& 69.5 		\\
Li 				& 53.4 	& 74.9  	& 86.5 	& 18.2 	& 57.9 	& 76.7 		\\
Kd-network 	& 45.1 	& 61.7  	& 81.4		& 24.1 	& 48.4 	& 72.6			\\
MVCNN 		& 61.2		& 73.4  	& 84.3 	& 41.6 	& 66.2 	& 79.3 		\\
GIFT 			& 66.1 	& 81.1  	& 88.9		& 42.3		& 73.0 	& 84.3			\\
TCL			& \textbf{67.9}     & \textbf{84.0}        & 89.5		& 43.9		& \textbf{78.3}		& \textbf{86.9}			\\
\hline
TCL(VGG) 			& 64.5       & 82.1        & 88.4		& 36.5		& 71.0		& 82.7			\\
Our & 67.4 & 83.6 & \textbf{89.7} 	& \textbf{46.1} & 75.4 & 85.8 \\
\bottomrule[2pt]
\end{tabular}
}
\caption{The performance (\%) comparison on SHREC16 perturbed dataset.}
\label{tab_shrec}
\end{table}

\noindent\textbf{Comparison with the state-of-the-arts.} 
On ModelNet40 dataset, we choose 3D ShapeNets~\cite{wu20153d}, DeepPano~\cite{shi2015deeppano}, MVCNN~\cite{su2015multi},
PANORAMA-NN~\cite{Sfikas2017Exploiting}, RED~\cite{Song2017Ensemble}, 
Siamese CNN-BiLSTM~\cite{dai2018siamese}, GVCNN~\cite{feng2018gvcnn}, TCL~\cite{he2018triplet} and GIFT~\cite{bai2016gift} methods for comparison. The experimental results and comparison among different methods are presented in Tab. \ref{tab_sr}. 
Our method ($L_{CIP}$ + center loss) achieves retrieval AUC of $88.21\%$ and MAP of $87.22\%$ which is the best among different methods. 
Besides, as the current state-of-the-art view-based method on ModelNet40, TCL is trained on VGG\_11 which has 3 more convolution layers than VGG\_M and adopts batch normalization \cite{DBLP:journals/corr/IoffeS15}. 
Compared with it, $L_{CIP}$ only loses $0.21\%$ of MAP on performance.
We also re-conduct TCL using VGG\_M according to the parameter settings in the paper (refer to Tab. \ref{tab_comploss}).  

For the evaluation in ShapeNetCore55 dataset, 
we carry out our method ($L_{CIP}$ + softmax loss) on the more challenging perturbed version.
For a fair comparison, we re-implement ``TCL + softmax loss'' on VGG\_M which is indicated as TCL(VGG) in Tab. \ref{tab_shrec}.
We also choose Wang~\cite{savva2016shrec}, Li~\cite{savva2016shrec}, K-d network~\cite{klokov2017escape}, MVCNN~\cite{su2015multi} and GIFT~\cite{bai2016gift} methods for comparison.
We take two types of results in the competition, namely macro and micro.
As shown in Tab. \ref{tab_shrec}, our method achieves the state-of-the-art performance.

\begin{table}
\begin{center}
\begin{tabular}{lccc}
\toprule[2pt]
Loss function 									& &AUC 				& MAP\\
\toprule[1.5pt]
softmax loss 									& & 81.16\% 			& 79.91\% \\
center loss + softmax loss  					& & 83.20\% 			& 82.04\% \\
triplet loss + softmax loss						& & 83.51\%			& 82.40\% \\
triplet-center loss(VGG)					& & 85.70\% 			& 84.64\% \\
coco loss(VGG)							& & 86.76\% 			& 85.72\% \\
\hline
$L_{cluster}$ + softmax loss    				& & 83.11\% 			& 81.94\% \\
$L_{ortho,batch}$  + softmax loss   			& & 87.16\% 			& 86.16\% \\
$L_{cluster}$ +  $L_{ortho,batch}$			& & 87.21\% 			& 86.17\% \\
$L_{CIP}$ ($L_{cluster}$ + $L_{ortho}$)		& & 87.56\% 			& 86.49\% \\
\hline
$L_{CIP}$ + softmax loss					& & 88.07\% 			& 87.08\%\\
$L_{CIP}$ + center loss						& & \textbf{88.21\%} 	& \textbf{87.22\%} \\
\bottomrule[2pt]
\end{tabular}
\end{center}
\caption{The performance comparison with different loss functions on ModelNet40.}
\label{tab_comploss}
\end{table}

\subsection{Comparison with other loss functions} 
\label{section_comploss}
To demonstrate the efficiency of our loss functions, we set various comparison experiments on ModelNet40 dataset.
Besides the commonly used loss functions,
we also carry out two well-known methods, coco loss and triplet-center loss, on VGG\_M.  
The experiment results are shown in Tab. \ref{tab_comploss}.
Our comparison contains 3 parts:

The first is the comparison with commonly used metric loss functions.
As a well-designed deep metric loss, $L_{CIP}$ dramatically improves the performance in MAP by $6.58\%$ from softmax loss supervision.
And our method outperforms TCL and coco loss which are improved efficient methods.
Moreover, for the comparison of $L_{ortho}$ and $L_{ortho,batch}$,
we can see that they achieve similar results when combining with $L_{cluster}$.
It also should be noted that the optimization will diverge if the CNN is trained solely with  $L_{cluster}$ or $L_{ortho}$ because both two terms are necessary.

The second part explores the combination with other loss functions.
Both of our loss functions can combine softmax loss separately since softmax loss already has pull term and push term in loss design, and the results show that the combinations bring a significant improvement.
Compared with center loss, $L_{cluster}$ achieves $81.94\%$ in MAP when jointly trained with softmax loss, which is evidence that $L_{cluster}$ has a comparable polymerization ability.
And to prove the efficiency of $L_{ortho}$, we conduct ``$L_{ortho}$ + softmax loss''.
This combination brings a large improvement of $5.31\%$ in MAP compared with single softmax loss, which means $L_{ortho}$ has a powerful capacity in enlarging inter-class distance.

Finally, we train the CNN with three loss functions and obtain an impressive performance. 
``$L_{CIP}$ + softmax loss'' achieves a MAP of $87.08\%$ and ``$L_{CIP}$ + center loss'' achieves a MAP $87.22\%$ on Modelnet40.
Note that the loss weight of softmax loss and center loss is 0.1 and 0.0003 respectively in the combination with our loss function.
All these experiments demonstrate our loss design can generate robust and discriminative features.

\subsection{Discussion}
\begin{figure}
\centering
\includegraphics[width=0.9\linewidth]{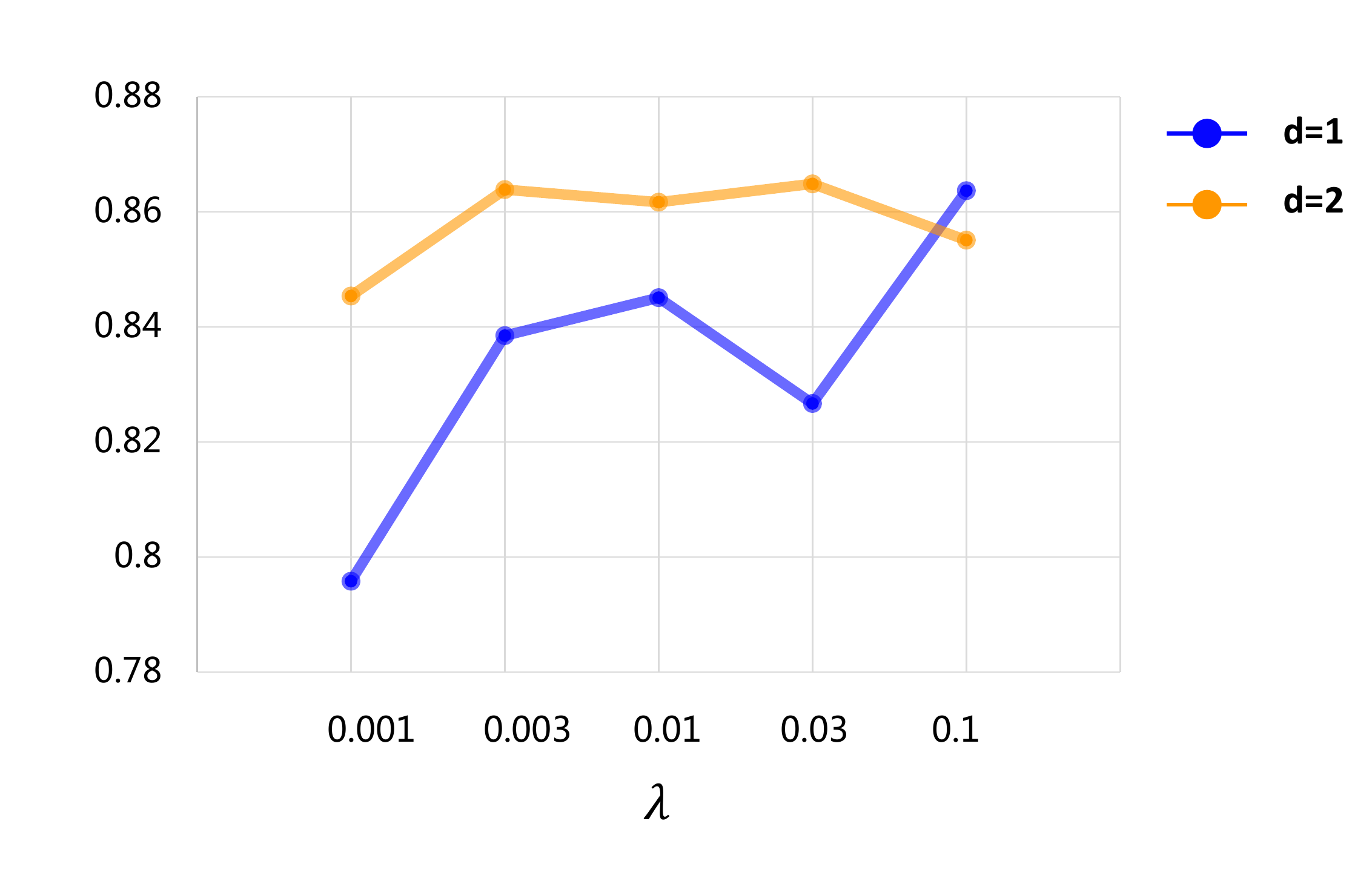}
\caption{The retrieval performance (MAP) of different hyper-parameter settings on ModelNet40.}
\label{fig_lambda}
\end{figure}

\noindent\textbf{Sensitiveness of hyper-parameter.} 
In Eq. \ref{equation_cluster}, $d$ is a constant value for stability. 
In our method, we choose $d=2$ because this choice can reduce the sensitiveness of trade-off parameter $\lambda$. 
To investigate the influence of $d$ and $\lambda$ on the performance, we conduct experiments supervised by CIP loss on ModelNet40.
The experiment results are shown in Fig. \ref{fig_lambda}.
We can see that CIP loss can obtain state-of-the-art results with different values of $d$ by adjusting $\lambda$, but the sensitiveness of $\lambda$ is different.
The performance varies a lot when  $d=1$.
With $d=2$, CIP loss can converge stably with a wide range of $\lambda$, which means our loss design guarantees stability in the optimization.

\noindent\textbf{Visualization of experiment results.} 
We use t-SNE \cite{ljp2008tsnes} to visualize the experiment results with different loss functions in Fig. \ref{fig_sne}.
From the figure, we can see that under the supervision of CIP loss, the distance between classes is increased compared with softmax loss and center loss.
Compared with TCL and coco loss, the intra-class variance is further decreased with CIP loss and the geometric separation become more clear. 

\section{Conclusion}
In this paper, we propose a novel loss function named Collaborative Inner Product Loss with regard to large-scale 3D shape retrieval task.
The Inner product is employed in the loss design which not only imposes a more strict constraint but also guarantees stability in optimization.
The proposed loss function consists of Cluster loss and Ortho loss that play different roles: one for reducing the intra-class distance and the other for enlarging the inter-class margin, and both of them can combine with other commonly used loss functions.
Experimental results on two large-scale datasets have proven the superiority of our loss functions.

\section{Acknowledgements} 
This work is supported by the Beijing Municipal Natural Science Foundation  (No.L182014), and the Open Project Program of State Key Laboratory of Virtual  Reality Technology and Systems, Beihang University (No.VRLAB2019C05).


\bibliographystyle{named}
\bibliography{ijcai19}

\end{document}